\documentclass[letterpaper, 10 pt, conference]{ieeeconf}  

\IEEEoverridecommandlockouts                              

\usepackage{graphics} 
\usepackage{graphicx}
\usepackage{algorithmic}
\usepackage{amsmath} 
\usepackage{amssymb}  
\usepackage{cite}
\usepackage{amsfonts}   
\usepackage{booktabs}  
\usepackage{CJKutf8}
\usepackage{xcolor}
\usepackage{float}
\usepackage{censor}
\usepackage{multirow}
\usepackage[ruled,vlined]{algorithm2e}

\usepackage{booktabs}

\usepackage{multirow}

\usepackage{makecell}

\title{\LARGE \bf Keyframe-Guided Structured Rewards for Reinforcement Learning in Long-Horizon Laboratory Robotics}

\author{
  Yibo Qiu\textsuperscript{$\dagger$}, 
  Shu'ang Sun\textsuperscript{$\dagger$}, 
  Haoliang Ye\textsuperscript{$\dagger$}, 
  Ronald X Xu\textsuperscript{$\ast$}, 
  Mingzhai Sun\textsuperscript{$\ast$}
  \thanks{$^\dagger$Equal contribution. $^\ast$Corresponding authors.}
  \thanks{All authors are with the Suzhou Institute for Advanced Research, University of Science and Technology of China, Suzhou, Jiangsu, China, and also with the School of Biomedical Engineering, Division of Life Sciences and Medicine, University of Science and Technology of China, Hefei, Anhui, China.}
  \thanks{Emails:\{alexandreqiu,sunshuang\}@mail.ustc.edu.cn, 15005704869@163.com (H. Ye), \{xux, mingzhai\}@ustc.edu.cn.}
  \thanks{This work was supported by the Gusu Leading Talent Entrepreneurship and Innovation Program (Grant No. ZL2024349).}
}

\begin{document}
\begin{CJK*}{UTF8}{gbsn}

\maketitle
\thispagestyle{empty}
\pagestyle{empty}

\begin{abstract}
Long-horizon precision manipulation in laboratory automation, such as pipette tip attachment and liquid transfer, requires policies that respect strict procedural logic while operating in continuous, high-dimensional state spaces. However, existing approaches struggle with reward sparsity, multi-stage structural constraints, and noisy or imperfect demonstrations, leading to inefficient exploration and unstable convergence. 
We propose a \textbf{Keyframe-Guided Reward Generation Framework} that automatically extracts kinematics-aware keyframes from demonstrations, generates stage-wise targets via a diffusion-based predictor in latent space, and constructs a geometric progress-based reward to guide online reinforcement learning. The framework integrates multi-view visual encoding, latent similarity-based progress tracking, and human-in-the-loop reinforcement fine-tuning on a Vision-Language-Action backbone to align policy optimization with the intrinsic stepwise logic of biological protocols. 
Across four real-world laboratory tasks, including high-precision pipette attachment and dynamic liquid transfer, our method achieves an average success rate of \textbf{82\%} after 40--60 minutes of online fine-tuning. Compared with HG-DAgger (42\%) and Hil-ConRFT (47\%), our approach demonstrates the effectiveness of structured keyframe-guided rewards in overcoming exploration bottlenecks and providing a scalable solution for high-precision, long-horizon robotic laboratory automation.

\end{abstract}

\section{INTRODUCTION}
\begin{figure*}[htbp]
  \centering  
  \includegraphics[width=1\linewidth]{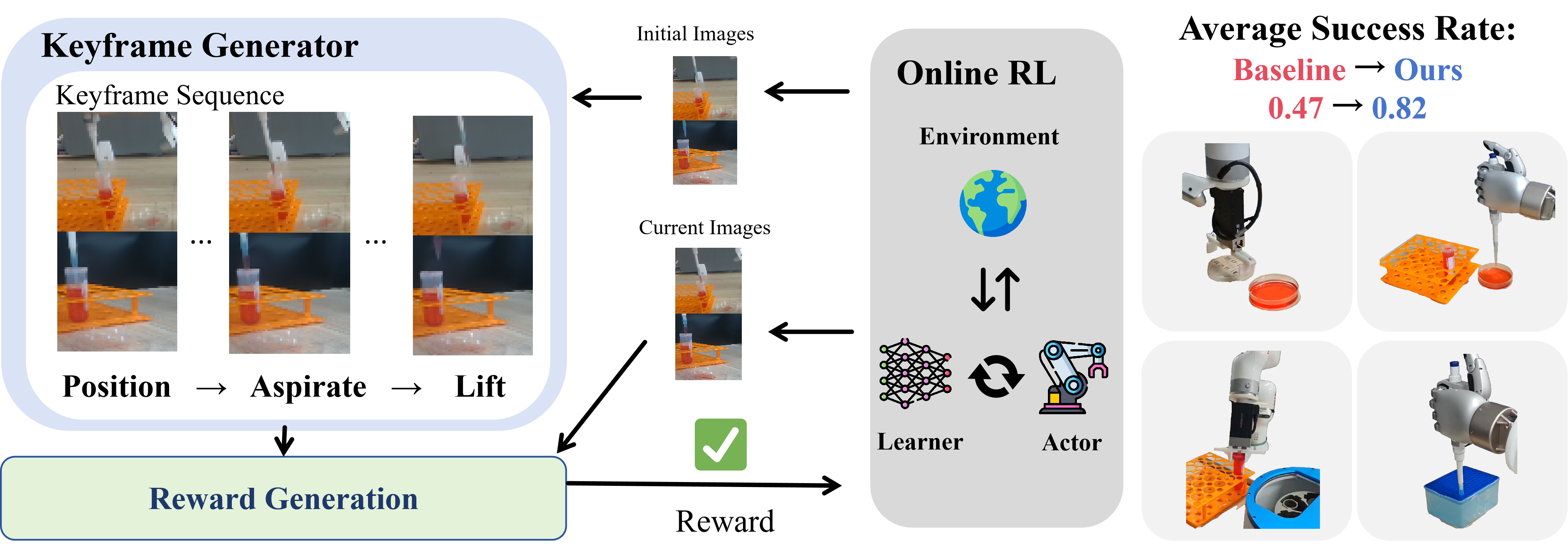}
  
  \vspace{-0.4cm}
  \caption{\textbf{Illustration of our Keyframe-guided Online RL framework.} Given an initial observation, the framework generates a sequence of multi-view keyframes (e.g., Position $\rightarrow$ Aspirate $\rightarrow$ Lift) to represent the target task. These keyframes act as intermediate goals to provide stage-wise rewards via state-similarity calculation. Compared to state-of-the-art baselines, our approach significantly accelerates convergence and achieves a superior average success rate of 82\%, outperforming the baseline by 47\%.}
  \label{fig:Illustration_motivation} 
  
  \vspace{-14pt}
\end{figure*}

For decades, researchers have pursued intelligent laboratory automation to enable robots to autonomously perform delicate operations such as pipetting and sample loading \cite{omair2023recent,angers2025roboculture}. Reliable execution of these procedures is essential for long-term biological experiments and for transitioning laboratory systems from rigid preprogrammed pipelines to flexible autonomous intelligence. Despite recent advances in robotic reinforcement learning (RL) and imitation learning (IL), achieving robust, high-precision manipulation under strict stepwise procedural constraints remains challenging.

A central difficulty arises from the long-horizon and multi-stage nature of laboratory tasks. For example, a complete pipetting procedure involves tightly coupled steps—alignment, insertion, aspiration, transfer, and reset—each governed by strict physical and logical constraints. However, standard RL formulations typically provide only sparse terminal rewards upon task completion. This creates a prolonged feedback vacuum across hundreds of intermediate transitions, resulting in inefficient exploration, unstable convergence in high-dimensional continuous action spaces, and insufficient guidance for precision-critical operations.

To mitigate these issues, prior work has explored imitation learning and human-in-the-loop reinforcement learning. HG-DAgger relies on behavior cloning from expert demonstrations \cite{kelly2019hg}, while Hil-SERL and Hil-ConRFT incorporate human intervention to improve safety and sample efficiency \cite{luo2025precise,chen2025conrft,luo2024serl}. Nevertheless, behavior cloning depends heavily on high-quality demonstrations and is sensitive to suboptimal data. Human-in-the-loop approaches improve safety but still rely on sparse or manually designed reward signals and do not explicitly exploit the inherent stage-wise structure of laboratory procedures. Furthermore, single-view perception introduces depth ambiguity and occlusion, increasing uncertainty in state estimation and reward computation.

This paper addresses the following question: \emph{How can we leverage the intrinsic stepwise structure of biological protocols to alleviate reward sparsity and improve learning efficiency without requiring carefully engineered rewards or expert-quality demonstrations?}

Inspired by the multi-stage nature of laboratory procedures, we propose a \textbf{Keyframe-Guided Reward Generation Framework} for long-horizon precision manipulation. The core idea is to automatically extract structurally meaningful keyframes from demonstrations using kinematics-aware latent dynamics analysis, and to transform these keyframes into stage-wise supervision signals for reinforcement learning.

Specifically, we introduce a coarse-to-fine keyframe extraction mechanism that detects critical operational nodes (e.g., positioning $\rightarrow$ aspiration $\rightarrow$ lifting) from non-expert demonstrations by analyzing latent feature dynamics. We further employ multi-view visual encoding and latent similarity metrics to construct a compact state representation, from which a geometric progress-based reward is generated to provide continuous intermediate guidance. This structured reward design aligns policy optimization with the procedural logic of laboratory workflows.

We validate the proposed framework on four real-world laboratory manipulation tasks involving long-horizon operations and sub-millimeter precision. Experimental results demonstrate substantial improvements in convergence speed and task success rate compared to state-of-the-art hybrid learning baselines. These findings suggest that structurally grounded keyframe rewards offer an effective and scalable solution for deploying reinforcement learning in precision laboratory automation.

The main contributions of this work are:
\begin{itemize}
    \item A kinematics-aware keyframe extraction mechanism that automatically identifies structurally critical states from non-expert demonstrations.
    \item A latent similarity-based stage-wise reward formulation that transforms extracted keyframes into progress-aware dense rewards for long-horizon RL.
    \item Integration of multi-view perception, keyframe-guided rewards, and human-in-the-loop fine-tuning on a Vision-Language-Action backbone for real-world laboratory manipulation.
    \item Extensive real-world validation across four laboratory automation tasks demonstrating significant improvements in convergence speed and success rate.
\end{itemize}

\section{RELATED WORK}

\subsection{Laboratory Automation and Robotic Manipulation}

Laboratory automation has become a cornerstone of modern life sciences and drug discovery, enabling high-throughput and high-precision experimental workflows \cite{qiu2025biomars, thurow2023strategies}. Recent systems have evolved from task-specific liquid-handling platforms \cite{torres2022automated} to modular cell culture systems and autonomous robots for synthetic chemistry \cite{hamm2024modular, dai2024autonomous, lu2024automated}. Despite these advances, most platforms remain highly specialized and rely on predefined procedural scripts, limiting their adaptability to diverse and evolving laboratory requirements \cite{wolf2025integration}.

Autonomous execution of complex laboratory protocols requires reliable coordination across multiple tightly coupled stages, often under sub-millimeter precision constraints \cite{zhao2025autonomous, seifrid2022autonomous, salazar2026adept}. While deep reinforcement learning (DRL) provides a promising framework for adaptive control, laboratory tasks are inherently long-horizon and multi-stage, making naive RL formulations vulnerable to sparse rewards and inefficient exploration in high-dimensional state spaces. Existing approaches typically do not explicitly leverage the intrinsic procedural structure of laboratory workflows when designing learning objectives.

\subsection{VLA Models and Reinforcement Fine-Tuning}

Vision-Language-Action (VLA) models have recently emerged as a scalable paradigm for generalizable robotic manipulation \cite{zhang2025pure,brohan2022rt,kim2024openvla,black2024pi_0,intelligence2025pi}. To adapt large-scale models to specific tasks, recent work integrates reinforcement learning and fine-tuning strategies, including offline RL \cite{huang2025co,zhang2025reinbot} and online RL refinement \cite{guo2025improving,xu2024rldg,jin2025dual}. These approaches demonstrate improved performance in multi-step manipulation scenarios.

However, the effectiveness of RL-based fine-tuning critically depends on reward design, particularly in sparse-reward environments \cite{ibrahim2024comprehensive, he2023robotic}. Physics-informed reward shaping \cite{fareh2025physics} and goal-conditioned RL \cite{he2023robotic} have been proposed to guide exploration, yet they typically require manually engineered reward terms or predefined subgoals. In laboratory automation settings, where procedural steps follow strict physical and logical constraints, designing such reward functions becomes increasingly challenging and task-specific \cite{deng2025reward}. Recent advancements in large models offer a promising alternative for automating reward engineering, allowing for the dynamic generation of task-specific rewards without requiring manual intervention or predefined subgoals\cite{yu2023language,wang2023robogen,ma2023eureka,rocamonde2023vision,chen2025tevir}.

In contrast to prior reward engineering approaches, we propose to automatically extract structurally meaningful keyframes from demonstrations and transform them into stage-wise progress-aware rewards. This allows reinforcement learning to align with the intrinsic stepwise structure of laboratory protocols without manual reward specification.

\section{METHOD}
\begin{figure*}[htbp]
  \centering  
  \includegraphics[width=1\linewidth]{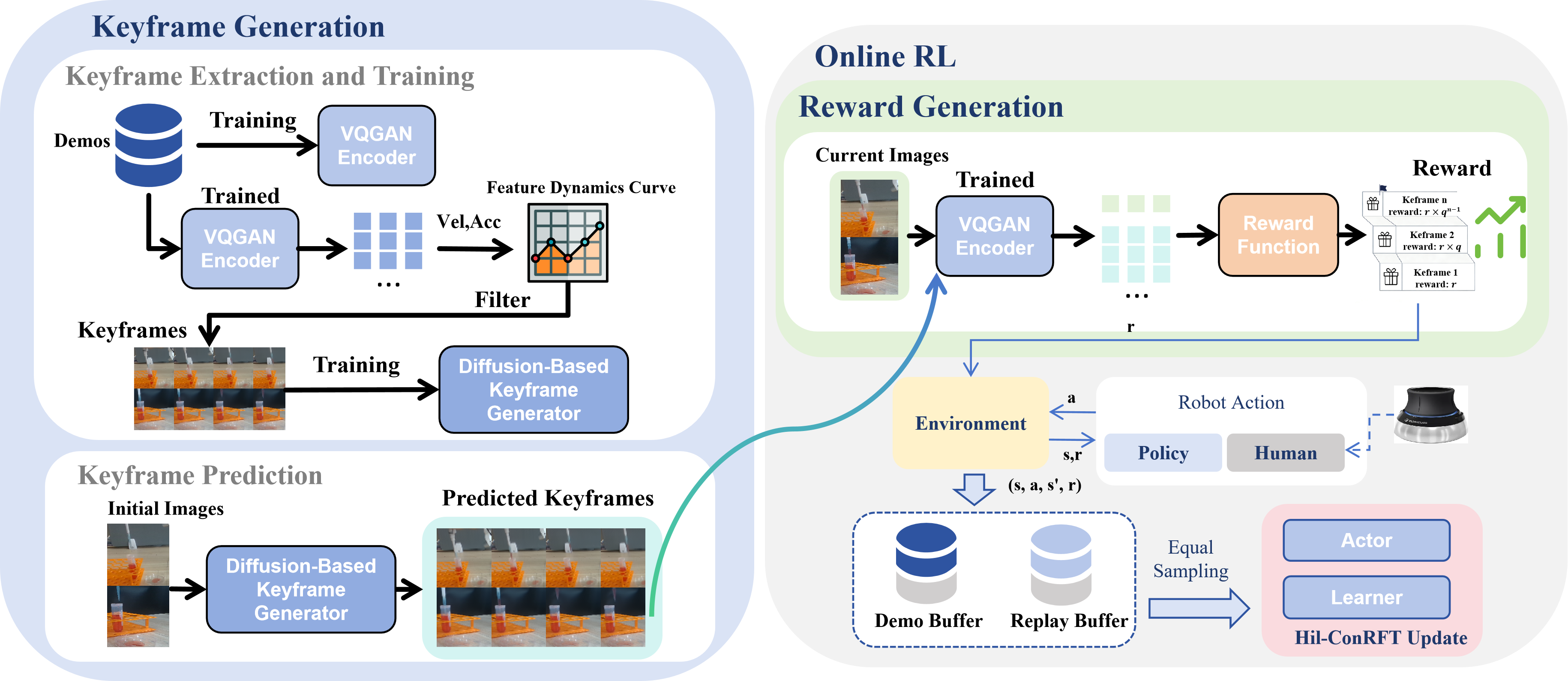}
  
  \vspace{-0.4cm}
  \caption{\textbf{Overview of our Keyframe-guided RL framework.} First, the system extracts keyframes by filtering feature dynamics from demonstrations. A diffusion-based generator is then trained to predict these keyframe sequences from initial observations. During online RL, the reward module calculates latent similarity to provide stage-wise guidance. These rewards drive the Hil-ConRFT update for policy learning.}
  \label{fig:Overview} 
  
  \vspace{-14pt}
\end{figure*}

This paper presents a keyframe-guided online reinforcement learning (RL) framework designed for laboratory automation. By integrating online interaction with multi-stage reward signals, the proposed system accelerates policy convergence and enables high-precision manipulation. As illustrated in Fig.~\ref{fig:Overview}, the framework consists of three primary modules: Latent Dynamics-Based Keyframe Extraction, Generative Diffusion-Based Reward Design, and Human-in-the-Loop (HiL) Policy Optimization.


\subsection{Latent Dynamics-Based Keyframe Extraction}
Keyframes are extracted from expert demonstration sequences. Given multi-view observations
$\mathbf{o}_t=\{\mathbf{o}_t^{\text{front}},\mathbf{o}_t^{\text{wrist}}\}$,
we use a discrete vector-quantized encoder $E$ pretrained on offline demonstrations to produce compact latents\cite{esser2021taming}.
We define a composite feature
\begin{equation}
\mathbf{f}_t=\text{concat}\!\left(E(\mathbf{o}_t^{\text{front}}),\ \lambda E(\mathbf{o}_t^{\text{wrist}})\right),
\end{equation}
with view weight $\lambda=1.5$.

We apply 1D Gaussian smoothing along the temporal axis to obtain $\tilde{\mathbf{f}}_t$, and compute latent velocity and acceleration as:
\begin{equation}
v_t=\|\tilde{\mathbf{f}}_t-\tilde{\mathbf{f}}_{t-1}\|_2,\qquad
a_t = v_t - v_{t-1}.
\end{equation}

\paragraph{Stage 1: Turning-Point Mining.}
Turning points are identified by detecting peaks in both $v_t$ and $|a_t|$,which results in a candidate set $\mathcal{K}$ of turning points.

\paragraph{Stage 2: Fixed-Budget Selection.}
From the candidate set $\mathcal{K}$, a fixed number of keyframes ($K=6$) are selected based on the transition score $s(k)=\max{v_k,|a_k|}$. Each selected keyframe produces view-specific latents:
\begin{equation}
\bar{\mathbf{z}}_h^v = E\!\left(\mathbf{o}_{k_h}^v\right).
\end{equation}

\subsection{Generative Diffusion-Based Online Reward System}
During online execution, a diffusion-based model \cite{chen2025tevir} is used to generate a stage-wise target sequence:
\begin{equation}
\{\hat{\mathbf{z}}_h^v\}_{h=1}^{H},
\end{equation}
conditioned on the task instruction and initial observation. These generated targets align with the task stages, providing a generalized reference for the reward module.

\subsubsection{Anchor-Centric Sequence Construction}
To bridge the gap between extracted keyframes and the required training sequence length ($H=8$), we adopt an anchor-centric sampling strategy. The detected keyframes $\{k_h\}_{h=1}^{K}$ serve as fixed temporal anchors. The remaining slots are filled by uniformly sampling from the non-keyframe pool $\mathcal{P} = \{0,\dots,T-1\}\setminus\{k_h\}_{h=1}^K$, ensuring the final task state is included:
\begin{equation}
\begin{split}
S = & \{k_h\}_{h=1}^{K} \cup \{ \text{uniformly sampled} \\
    & \text{non-keyframe indices from } \mathcal{P} \}.
\end{split}
\end{equation}
This subset $\{\mathbf{o}_t\}_{t\in S}$ prioritizes key interaction nodes for reward calculation.

\subsubsection{Multi-View Fusion Similarity}
The alignment between the current state and the target sequence is evaluated via cosine similarity in the latent space:
\begin{equation}
S_t(h) = \sum_{v \in \mathcal{V}} w_v \cdot 
\frac{ \mathbf{z}_t^v \cdot \hat{\mathbf{z}}_h^v }
{ \|\mathbf{z}_t^v\|_2 \, \|\hat{\mathbf{z}}_h^v\|_2 },
\end{equation}
where $\mathcal{V}=\{\text{front},\text{wrist}\}$ and $w_v$ are view weights (detailed in Appendix~\ref{app:hyperparams}).

\subsubsection{Progress Tracking and Reward Calculation}
We maintain a progress state $M_t\in\{1,\dots,H\}$ and advance it when
$S_t(M_t)\ge\theta$:
\begin{equation}
M_{t+1}=
\begin{cases}
M_t+1, & \text{if } S_t(M_t)\ge\theta \text{ and } M_t\le H\\
M_t, & \text{otherwise}.
\end{cases}
\end{equation}

The guided reward is formulated as
\begin{equation}
R_{\text{total}} = \epsilon_{\text{step}} + \mathbb{I}(M_{t+1}>M_t)\cdot G_{\text{stage}}(M_{t+1}),
\end{equation}
where $\epsilon_{\text{step}}=-0.05$ is a small step penalty. We use a geometric progression
\begin{equation}
G_{\text{stage}}(n) = r_1 q^{\,n-1}, \quad n=1,\dots,H,
\end{equation}
with a scaling factor \(q = 1.3\), and a fixed total reward budget
\begin{equation}
\sum_{n=1}^{H} G_{\text{stage}}(n) = 10.
\end{equation}

\subsection{Online Policy Evolution and Q-Learning Update}
Our policy backbone is based on Octo \cite{team2024octo}, a pre-trained Vision-Language-Action (VLA) model. The action head uses a diffusion policy to generate control signals through iterative denoising. For real-time online learning, we apply a consistency policy \cite{chen2023boosting,song2023consistency,karras2022elucidating} to distill the denoising process into a single-step inference. Stochastic exploration is maintained via noise sampling:
\begin{equation}
w \sim \mathcal{N}(0,\sigma^2\mathbf{I}),
\end{equation}
where $\sigma$ controls the initial noise scale. Different samples of $w$ yield diverse actions, facilitating better exploration.

The policy is updated using a mixed-buffer strategy, sampling equally from the online replay buffer $\mathcal{D}_{\text{online}}$ and the demonstration buffer $\mathcal{D}_{\text{demo}}$. We integrate a Human-in-the-Loop (HiL) mechanism where operators provide corrective interventions during unsafe or suboptimal behaviors; these corrections are stored in $\mathcal{D}_{\text{demo}}$ to guide the update. Our optimization follows the HiL-ConRFT framework \cite{chen2025conrft}.

\subsubsection{Critic Network Update}
The critic is optimized via Bellman regression with a target ensemble to reduce overestimation:
\begin{equation}
a' \sim \pi(\cdot|s'),  
y = r + \gamma \cdot \min_i Q_{\bar{\theta}_i}(s', a').
\end{equation}
where $\gamma$ is the discount factor
The critic loss is
\begin{equation}
\mathcal{L}_{\text{critic}} = \mathbb{E}_{(s,a,r,s') \sim \mathcal{D}}
\left[\left(Q_{\theta}(s,a) - y\right)^2\right],
\end{equation}
where $\mathcal{D} = \mathcal{D}_{\text{demo}} \cup \mathcal{D}_{\text{online}}$.

\subsubsection{Actor Network Update}
The actor is optimized by a weighted combination of behavior cloning and Q-maximization:
\begin{equation}
\mathcal{L}_{\text{actor}} = \lambda_{\text{bc}}\mathcal{L}_{\text{BC}} + \lambda_{\text{q}}\mathcal{L}_{\text{Q}},
\end{equation}
where $\lambda_{\text{bc}}=0.5$ and $\lambda_{\text{q}}=1$ .
The behavior cloning loss is
\begin{equation}
\mathcal{L}_{\text{BC}} = \mathbb{E}_{(s,a)\sim \mathcal{D}}\!\left[\|a-\pi_\phi(s)\|_2^2\right],
\end{equation}
and the Q-loss maximizes the expected critic value
\begin{equation}
\mathcal{L}_{\text{Q}} = -\mathbb{E}_{s\sim \mathcal{D}}\!\left[Q_{\theta}(s,\pi_\phi(s))\right].
\end{equation}

\begin{figure*}[htbp]
  \centering  
  \includegraphics[width=1\linewidth]{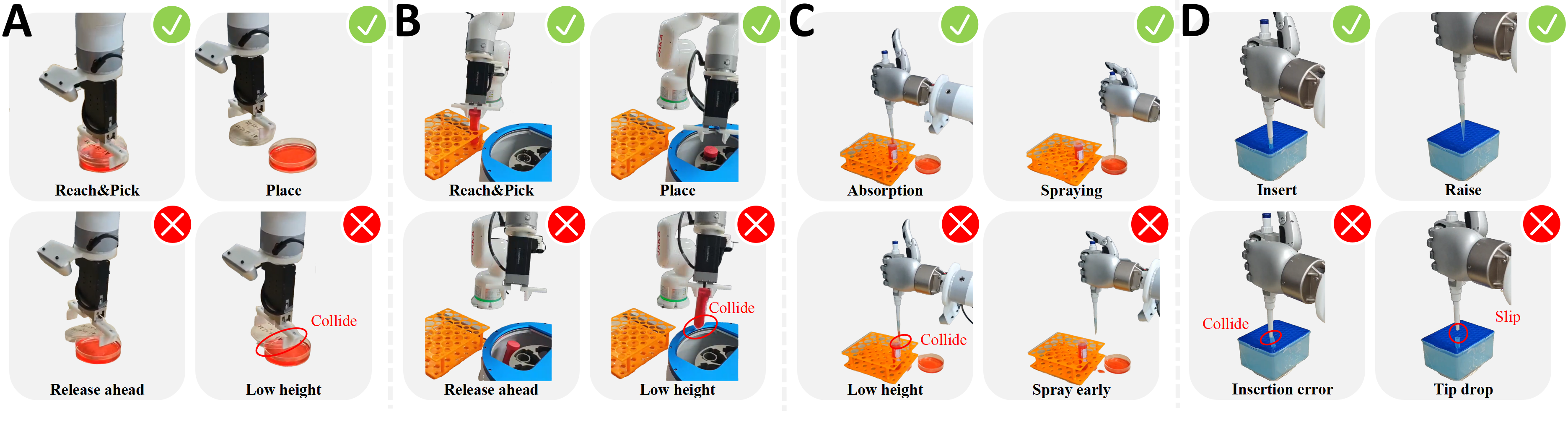}
  
  \vspace{-0.4cm}
  \caption{\textbf{Biological Laboratory task illustrations and common failure modes.} (A) Petri Dish De-lidding. This task involves using a gripper to lift and relocate the lid of a petri dish to the side. Common failures include premature release, or colliding with the dish base due to insufficient height. (B) Centrifuge Tube Loading. This task involves picking a tube from a rack and placing it into a centrifuge slot. Common failures include missing the tube, dropping it early, or colliding with the centrifuge rim. (C) Precision Liquid Transfer. This task involves absorbing liquid from a tube and spraying it into a petri dish. Common failures include misalignment with the tube opening, colliding with the tube wall, or spraying liquid before reaching the target. (D) Pipette Tip Attachment. This task involves inserting the pipette into a tip box to secure a new tip. Common failures include misalignment, insertion errors, or the tip slipping off during raising.}
  \label{fig:Task_illustrations_and_failure_modes} 

  
  \vspace{-14pt}
\end{figure*} \

\section{EXPERIMENTS}

\subsection{Experimental Platform and Task Parameter Settings}

To evaluate the effectiveness of the proposed keyframe-guided reward framework in real-world laboratory automation, we design four representative manipulation tasks (Fig.~\ref{fig:Overview}, Fig.~\ref{fig:Task_illustrations_and_failure_modes}). These tasks span object relocation (centrifuge tube loading and petri dish de-lidding), high-precision contact-rich manipulation (pipette tip attachment), and dynamic liquid handling (precision liquid transfer). All training and inference were conducted on a workstation equipped with an NVIDIA RTX 5880 Ada GPU. 

\textbf{Observation and Action Spaces.} 
All tasks are operated at a control frequency of 10 Hz. The observation space $\mathbf{s}_t$ consists of dual-view RGB images (a $128 \times 128$ wrist camera and a $256 \times 256$ side camera) and proprioceptive states, including end-effector (EE) poses and gripper/pipette status. The action space $\mathbf{a}_t$ is defined as a 6-DoF incremental EE pose in Euclidean space, augmented by task-specific dimensions: a 1D binary gripper state for Tasks 1 \& 2, and coordinated dexterous hand/pipette triggers for Tasks 3 \& 4.

\textbf{Task Specifications and Randomization.} 
We evaluate our framework on four representative laboratory automation tasks using a Jaka Minicobo robot:
\textit{Tasks 1 \& 2 (Object Relocation):} These involve Petri Dish De-lidding and Centrifuge Tube Loading. A parallel gripper is used to execute multi-stage pick-and-place workflows.
\textit{Task 3 (Pipette Tip Attachment):} A high-precision, contact-rich task requiring sub-millimeter accuracy, utilizing a Linkerhand O7 dexterous hand to interface with pipette tips.
\textit{Task 4 (Precision Liquid Transfer):} A complex dynamic task requiring synchronized control of the dexterous hand and pipette triggers for aspiration and dispensing.
To ensure policy robustness, initial object positions are randomized within a $3 \times 3 \text{ cm}^2$ ($2 \times 2 \text{ cm}^2$ for Task 3) horizontal area across all episodes.

\subsection{Comparison with Baseline Methods}

\textbf{Selection of baseline methods.} To comprehensively evaluate the effectiveness of the proposed method in real-world laboratory scenarios, we compare our method (Ours) with mainstream policy fine-tuning and reinforcement learning baselines across four representative manipulation tasks shown in Fig.~\ref{fig:Task_illustrations_and_failure_modes}. The selected baseline methods primarily include: HG-DAgger, a classic method that achieves policy fine-tuning through supervised learning combined with manual online corrections; Hil-ConRFT, an advanced learning framework combining human interventions with reinforcement fine-tuning; and Hil-SERL , a reinforcement learning method that integrates human interventions and utilizes offline data to accelerate online exploration.

\textbf{Experimental Settings and Fairness.} To ensure a rigorous and unbiased comparison, we unified the underlying configurations across all evaluated methods, despite the inherent differences between laboratory manipulation scenarios and the original validation environments of certain baselines. For methods involving human-in-the-loop interventions, the frequency and criteria for human takeover and corrective data were strictly standardized during training. Regarding the reward structure, all baseline methods were implemented with a penalty-based dense reward: a step-wise penalty of $-0.05$ was assigned for each incomplete frame, followed by a terminal reward of $+10$ upon task completion. This configuration was specifically designed to align the total potential reward magnitude with our proposed framework. Furthermore, the replay buffer initialization and mixed sampling mechanisms were synchronized across all baselines to eliminate performance discrepancies stemming from engineering-level discrepancies.

\begin{figure*}[htbp]
  \centering  
  \includegraphics[width=1\linewidth]{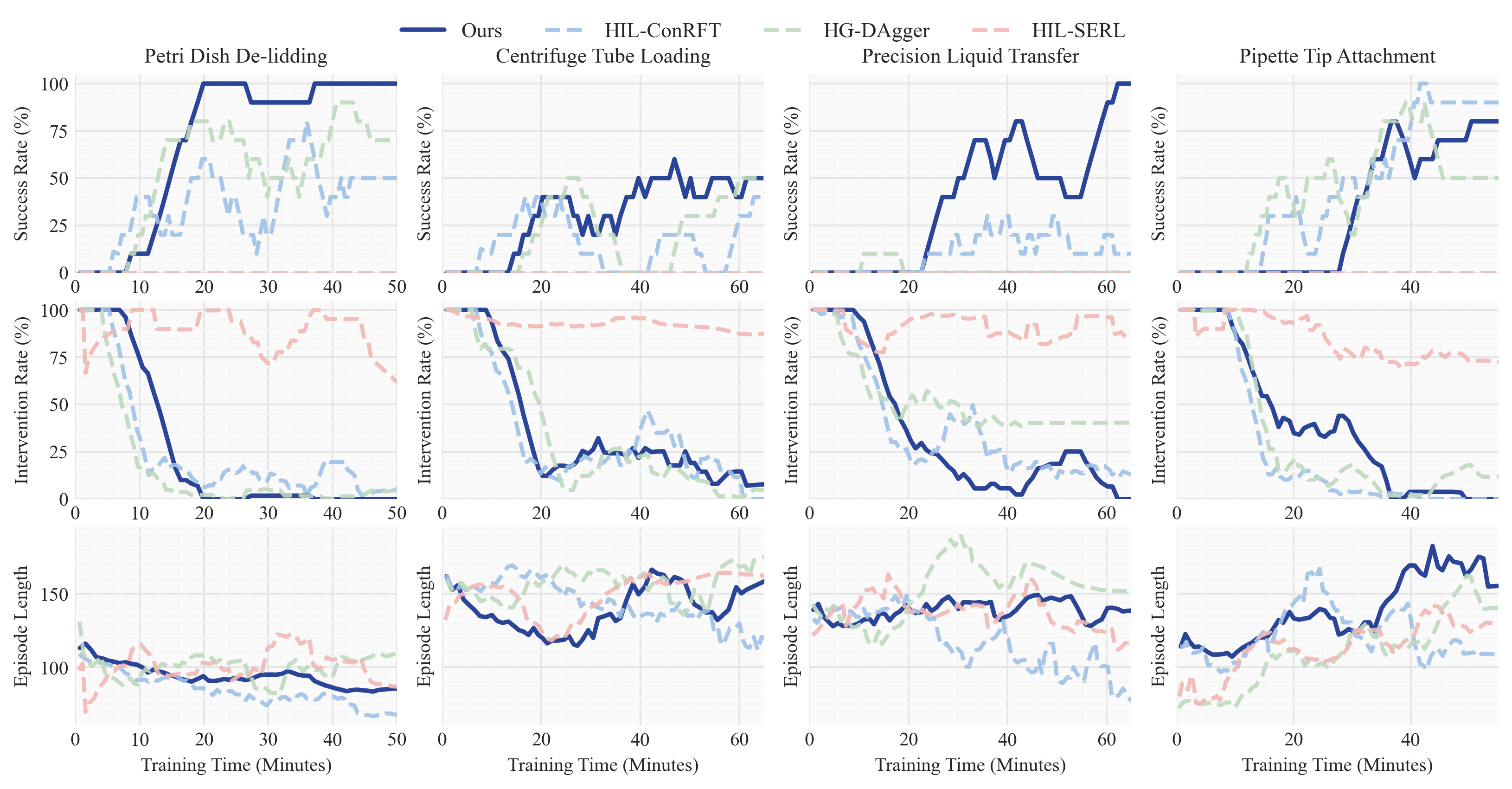}
  
  \vspace{-0.4cm}
  \caption{\textbf{Learning curves during online training.} This figure presents the success rates, intervention rates, and episode lengths for HIL-ConRFT, HG-DAgger, HIL-SERL, and our method across four representative laboratory tasks. The metrics are displayed as a running average over 10 episodes against training time in minutes.}
  \label{fig:Learning_curves} 
  
  \vspace{-14pt}
\end{figure*}

\textbf{Comprehensive performance comparison.} Table \ref{tab:performance} details the final evaluation results of each method across the four tasks, and Fig.~\ref{fig:Learning_curves} displays the corresponding online learning curves. Experimental results demonstrate that after approximately 40-60 minutes of real-world online fine-tuning, our keyframe-guided method achieves an overwhelming advantage, reaching an average success rate of 82\% across the four tasks. In contrast, the supervised fine-tuning method HG-DAgger, which relies heavily on data quality, achieves an average success rate of only 42\%; the reinforcement fine-tuning method Hil-ConRFT, lacking intermediate state guidance, yields only 47\%; and the reinforcement learning baseline Hil-SERL , relying on conventional dense or sparse rewards, falls into a severe exploration dilemma when facing these high-precision, long-horizon laboratory tasks, with a success rate of nearly 0\%. 






\begin{figure}[htbp]
  \centering  
  \includegraphics[width=1\linewidth]{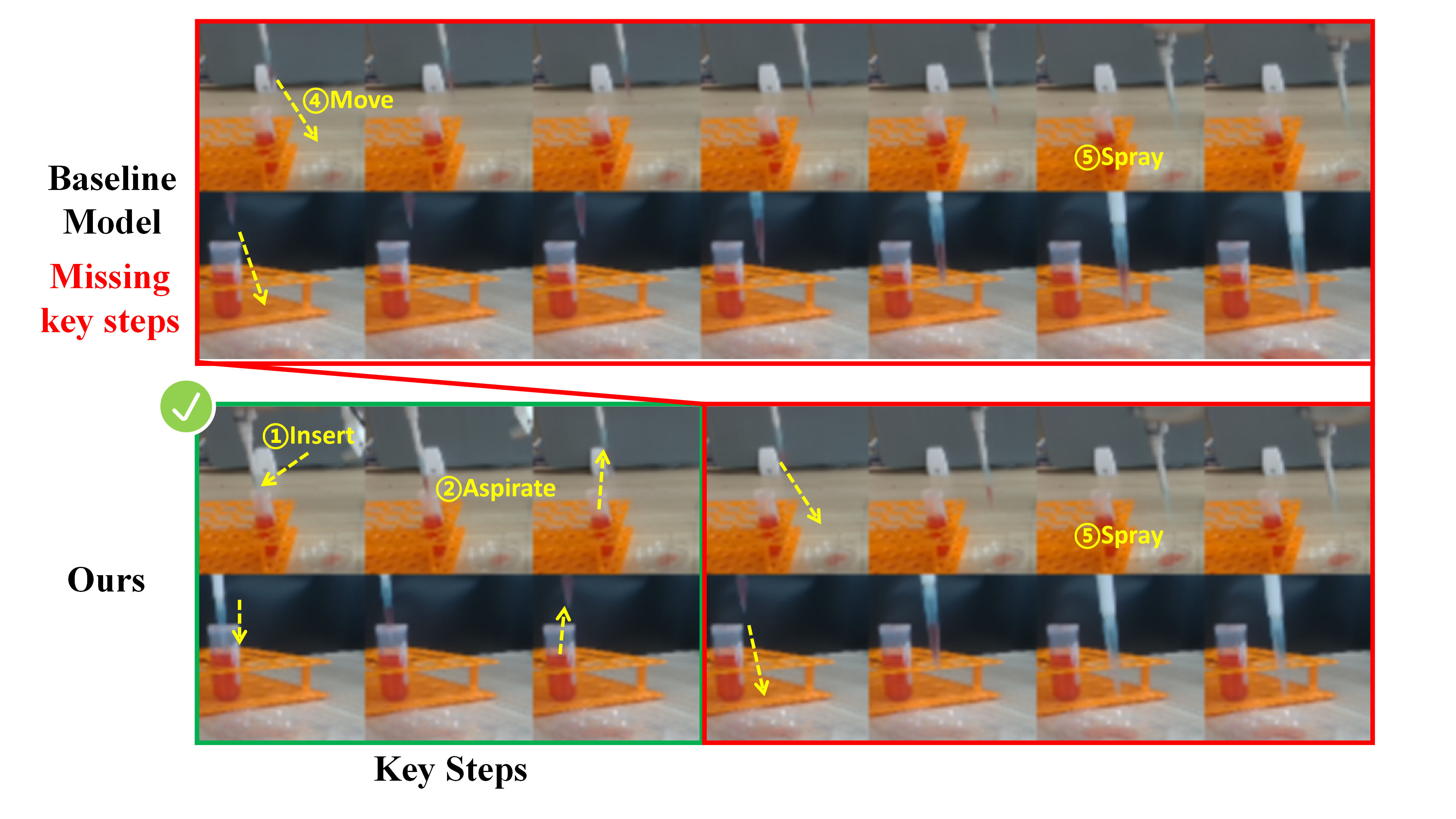}
  
  \vspace{-0.4cm}
  \caption{\textbf{Impact of keyframe introduction.} Precision liquid transfer requires a strict sequence from insertion to dispensing. Unlike our kinematics-heuristic extraction, the baseline uses uniform sampling and misses critical bottleneck states like insertion and aspiration. This leads to erroneous spatial guidance and execution failure. In contrast, our method anchors these intermediate states, ensuring correct step-wise logic.}
\label{fig:keyframe_impact}
  
  \vspace{-14pt}
\end{figure}



\begin{table}[tb]
  \centering
  \footnotesize  
  \caption{Performance Comparison of Different Methods on Robotic Manipulation Tasks}
  \label{tab:performance}
  \begin{tabular}{ccccc}
    \toprule  
    Task & Ours & \begin{tabular}[c]{@{}c@{}}HG-\\DAgger\end{tabular} & \begin{tabular}[c]{@{}c@{}}HIL-\\ConRFT\end{tabular} & \begin{tabular}[c]{@{}c@{}}HIL-\\SERL\end{tabular} \\
    \midrule  
    Petri Dish De-lidding            & \textbf{1.00} & 0.70 & 0.50 & 0.00 \\ 
    Centrifuge Tube Loading & \textbf{0.50} & \textbf{0.50} & 0.40 & 0.00 \\ 
    Precision Liquid Transfer      & \textbf{1.00} & 0.00 & 0.10 & 0.00 \\ 
    Pipette Tip Attachment          & 0.80 & 0.50 & \textbf{0.90} & 0.00 \\ 
    \cmidrule{1-5}     
    Mean                & \textbf{0.82} & 0.42 & 0.47 & 0.00 \\ 
    \bottomrule 
  \end{tabular}
  \vspace{-8pt}
\end{table}

\subsection{Results Analysis}

In this section, we provide an in-depth analysis of the performance of the proposed method (Ours) and baseline models in laboratory automation tasks, combining the success rate data in Table \ref{tab:performance}, the key step visualization in Fig.~\ref{fig:keyframe_impact}, and the online learning curves in Fig.~\ref{fig:Learning_curves}.

\textbf{Keyframe-guided dense rewards significantly improve multi-task success rates and generalization capabilities.} As shown in Table \ref{tab:performance}, the proposed keyframe-guided reinforcement learning framework achieves the best performance across four laboratory manipulation tasks, with an average success rate of 82\%. This is a substantial improvement compared to the supervised fine-tuning method HG-DAgger (42\%), which relies on manual corrections, and the reinforcement fine-tuning method Hil-ConRFT (47\%). Meanwhile, the reinforcement learning baseline Hil-SERL fails in these tasks, achieving a success rate of nearly 0\%. 

\textbf{The proposed method effectively overcomes structural defects in long-horizon and high-precision contact tasks.} Taking the "Precision Liquid Transfer" task as an example, the success rates of HG-DAgger and Hil-SERL are both 0\%, and Hil-ConRFT is only 10\%, whereas our method achieves a 100\% success rate. By analyzing the step-wise visualization in Fig.~\ref{fig:keyframe_impact}, the root cause becomes evident: the pipetting task involves strict physical logic constraints, including sequential steps such as inserting into the centrifuge tube to aspirate liquid, moving out of the tube, and moving above the petri dish to spray. Due to the lack of intermediate state guidance, baseline models often ignore intermediate key steps, resulting in severe collisions between the pipette tip and the centrifuge tube wall. In contrast, our method successfully captures these crucial intermediate steps by automatically extracting and generating a sequence of keyframes. This mechanism not only corrects the "skipping steps" defect of the baseline methods but also effectively ensures the safety and compliance of precision instrument operations.

\textbf{Extremely high sample efficiency and rapidly decreasing human intervention rate.} Observing the training curves in Fig.~\ref{fig:Learning_curves}, within the 40-60 minute training window, the success rate curve of Ours exhibits the steepest upward trend across all tasks. At the same time, while all methods require high human intervention in the early stages, the intervention rate of Ours drops rapidly to near zero in a short period. This demonstrates that the agent can swiftly absorb the intermediate guidance signals provided by the multi-view keyframes to establish correct state-action mappings. 

\begin{table}[htbp]
    \centering
    \scriptsize 
    \caption{Quantitative Analysis of Multi-Level Keyframe Identification Accuracy}
    \label{tab:keyframe_hierarchy_results}
    \setlength{\tabcolsep}{4pt} 
    \begin{tabular}{c c c c c} 
        \toprule
        \multirow{2}{*}{Task} & \multicolumn{2}{c}{\makecell{Step-wise \\ Coverage (All) $\uparrow$}} & \multicolumn{2}{c}{\makecell{Core Milestone \\ Recall (Core) $\uparrow$}} \\
        \cmidrule(lr){2-3} \cmidrule(lr){4-5}
        & w/o KE & w/ KE & w/o KE & w/ KE \\ 
        \midrule
        Petri Dish De-lidding      & 0.817 & \textbf{0.917} & 0.550 & \textbf{0.900} \\
        Centrifuge Tube Loading    & 0.667 & \textbf{0.933} & 0.350 & \textbf{0.850} \\
        Precision Liquid Transfer  & 0.550 & \textbf{0.900} & 0.400 & \textbf{1.000} \\
        Pipette Tip Attachment     & 0.933 & \textbf{0.983} & 0.800 & \textbf{1.000} \\
        \midrule
        \textbf{Mean}              & 0.742 & \textbf{0.933} & 0.525 & \textbf{0.938} \\
        \bottomrule
    \end{tabular}
    \vspace{-8pt}
\end{table}


\subsection{Ablation Studies}
\label{sec:ablation}

To systematically evaluate the contribution of our proposed modules, we conduct ablation studies focusing on the keyframe extraction mechanism and the keyframe-guided dense reward.

\textbf{Effectiveness of Keyframe Extraction Mechanism}
To isolate the effect of our kinematics-heuristic keyframe extraction (KE), we conduct an ablation study by replacing this module with a standard time-based uniform sampling strategy (denoted as \textit{w/o KE}). As reported in Table \ref{tab:keyframe_hierarchy_results}, we evaluate the performance using two metrics: \textbf{Step-wise Coverage (All)} for overall sequence alignment and \textbf{Core Milestone Recall (Core)} for the identification of critical operational nodes. 

Quantitatively, our KE method significantly outperforms the baseline, achieving an average Step-wise Coverage of $0.933$ compared to $0.742$. More importantly, the advantage is even more pronounced in the Core Milestone Recall, where our method maintains a high recall of $0.938$, while the baseline drops to $0.525$. This gap is most evident in the ``Precision Liquid Transfer'' task, where KE achieves a perfect recall ($1.000$) of core milestones, whereas uniform sampling frequently misses these transient but essential states ($0.400$). \textbf{Table \ref{tab:task_hyperparams}} provides the semantic grounding for these core milestones, detailing the specific physical transitions—such as tip-to-liquid contact and aspiration—that our mechanism successfully prioritizes. These results validate that capturing kinematics-aware keyframes is superior to uniform sampling for preserving the underlying logic of complex laboratory procedures.

Qualitatively, as visually compared in Fig.~\ref{fig:keyframe_impact}, the uniform sampling approach fails to capture the physical bottleneck states characterized by sudden kinematic changes. Critical intermediate steps, such as inserting the pipette tip into the centrifuge tube and aspirating the liquid, are completely omitted. Without these bottleneck states anchoring the trajectory, the generated sequence provides erroneous spatial guidance. This structural defect directly leads to critical failure modes during execution, such as colliding with the tube wall or misaligning the tip (as illustrated in Fig.~\ref{fig:Task_illustrations_and_failure_modes}C). This ablation robustly demonstrates that our keyframe extraction mechanism is indispensable for capturing the underlying physical logic of long-horizon tasks.

\textbf{Impact of Keyframe-Guided Reward.}
To isolate the contribution of our keyframe-guided reward generation, we compare our full pipeline with HIL-ConRFT. It is crucial to note that HIL-ConRFT shares the exact same VLA backbone (Octo), consistency policy action head, and human-in-the-loop training paradigm as our method. The fundamental difference is that HIL-ConRFT relies solely on standard sparse terminal rewards, completely ablating our stage-wise keyframe guidance. As quantitatively shown in Table \ref{tab:performance} and the learning curves in Fig.~\ref{fig:Learning_curves}, removing the keyframe-guided dense reward causes a severe performance drop. Most notably, in the complex "Precision Liquid Transfer" task, the success rate plummets from 100\% to a mere 10\%. Furthermore, without the intermediate state guidance provided by our generated keyframes, the baseline policy frequently falls into local optima, exhibiting erroneous behaviors such as premature spraying or insufficient lifting height (Fig.~\ref{fig:Task_illustrations_and_failure_modes}). 

\section{CONCLUSIONS}

This paper presented a keyframe-guided reinforcement learning framework for long-horizon precision manipulation in laboratory automation. By leveraging kinematics-aware keyframe extraction and diffusion-based stage target generation, the proposed method transforms demonstrations into structured, progress-aware reward signals without requiring manually engineered reward functions or expert-quality annotations. The resulting stage-wise reward mitigates temporal credit assignment challenges and aligns policy optimization with the intrinsic procedural logic of laboratory workflows.

Extensive real-world experiments across four multi-stage tasks demonstrate improved convergence speed, reduced reliance on human intervention, and an average success rate of 82\% under online fine-tuning. The framework effectively addresses common failure modes in sequential precision tasks, particularly in high-accuracy liquid handling scenarios.

While the current implementation relies on RGB-based latent similarity, which can be sensitive to visual artifacts such as reflections or transparency, future work will explore multimodal representations incorporating tactile sensing, depth perception, and 3D geometric features. We also aim to integrate automated environment reset mechanisms to further reduce human involvement and enhance scalability. Overall, this work provides a structured and scalable pathway for deploying reinforcement learning in real-world laboratory robotics.

\section*{APPENDIX}
\subsection{Task-Specific Configurations}
\label{app:hyperparams}

To balance global coordination and local precision, task-specific view weights $w_v$ and keyframe sequences are defined as follows:

\begin{table}[htbp]
    \centering
    \scriptsize
    \caption{Task-Specific View Weights and Keyframe Semantic Sequences}
    \setlength{\tabcolsep}{4pt}
    \begin{tabular}{p{2.2cm} c c c p{3.2cm}}
        \toprule
        Task & $w_{\text{wrist}}$ & $w_{\text{front}}$ & $\theta$ & Keyframe Sequence \\
        \midrule
        
        Petri Dish De-lidding
        & 0.5 & 0.5 & 0.7
        & Reach $\rightarrow$ \textcolor{red}{Grasp} $\rightarrow$ Lift $\rightarrow$ Relocate \\
        
        Centrifuge Tube Loading
        & 0.6 & 0.4 & 0.6
        & Approach $\rightarrow$ \textcolor{red}{Pick} $\rightarrow$ Align $\rightarrow$ Insert \\
        
        Precision Liquid Transfer
        & 0.7 & 0.3 & 0.6
        & Insert $\rightarrow$ \textcolor{red}{Aspirate} $\rightarrow$ Lift $\rightarrow$ Spray \\
        
        Pipette Tip Attachment
        & 0.7 & 0.3 & 0.8
        & Approach $\rightarrow$ \textcolor{red}{Insert} $\rightarrow$ Lift \\
        
        \bottomrule
    \end{tabular}

    \vspace{0.4em}
    {\footnotesize Note: Keyframes highlighted in \textcolor{red}{red} correspond to the core keyframes used for high-level semantic evaluation in Table~\ref{tab:keyframe_hierarchy_results}.}
    
    \label{tab:task_hyperparams}
    \vspace{-8pt}
\end{table}

\subsection{Algorithm}
\label{app:algorithm}

\begin{algorithm}[h!]
\small
\caption{Keyframe-Guided Online Policy Evolution}
\label{alg:framework}
\begin{algorithmic}[1]
\REQUIRE 
    $\mathcal{D}_{demo}, \mathcal{D}_{online}$: Demonstration and replay buffers; \\
    $\pi_{\phi}$: VLA policy with parameters $\phi$; \\
    $Q_{\theta}$: Critic network with parameters $\theta$; \\
    $E$: Visual encoder; $G$: Keyframe generator; \\
    $\theta_{sim}$: Similarity threshold; $H$: Total stages; \\
    $\lambda_{bc}, \lambda_{q}$: Balancing weights for BC and Q-learning.
\STATE Initialize stage index $M \leftarrow 0$ and $success \leftarrow \text{False}$.
\FOR{each episode}
    \STATE Obtain initial state $s_0$ and latent targets $\{\hat{z}_{h}^{v}\} \leftarrow E(G(s_0))$.
    \WHILE{NOT $success$ \AND NOT $human\_exit$}
        \STATE Sample action $a_t \sim \pi_{\phi}(\cdot|s_t)$.
        \IF{human intervention triggered}
            \STATE Execute expert $a_t^*$, store $(s_t, a_t^*)$ to $\mathcal{D}_{demo}$.
        \ELSE
            \STATE Execute $a_t$, observe $s_{t+1}$.
        \ENDIF
        
        \STATE // \textit{Latent Progress and Reward Calculation}
        \STATE Extract $z_{t+1}^v = E(s_{t+1}^v)$, compute similarity $S_t(M)$ via Eq. (6).
        \IF{$S_t(M) \ge \theta_{sim}$ \AND $M < H$}
            \STATE $r_{guided} \leftarrow G_{stage}(M)$, $M \leftarrow M + 1$.
        \ELSE
            \STATE $r_{guided} \leftarrow 0$.
        \ENDIF
        \STATE $r_t = \epsilon_{step} + r_{guided}$, store $(s_t, a_t, r_t, s_{t+1})$ to $\mathcal{D}_{online}$.
        
        \STATE // \textit{Policy ($\pi$) and Critic ($Q$) Update}
        \STATE Sample batch $\mathcal{B} \sim \{\mathcal{D}_{demo} \cup \mathcal{D}_{online}\}$.
        \STATE $y = r_t + \gamma \min_{i=1,2} Q_{\bar{\theta}_i}(s_{t+1}, \pi_{\phi}(s_{t+1}))$.
        \STATE Update $Q_{\theta}$ by minimizing Bellman error $\mathcal{L}_{critic}(\mathcal{B}, y)$.
        \STATE Update $\pi_{\phi}$ by minimizing $\lambda_{bc}\mathcal{L}_{BC}(\mathcal{B}) + \lambda_{q}\mathcal{L}_{Q}(\mathcal{B})$.
        
        \STATE $s_t \leftarrow s_{t+1}$, $t \leftarrow t + 1$.
        \STATE \textbf{if} $M = H+1$ \textbf{or} $human\_verified$ \textbf{then} $success \leftarrow \text{True}$.
    \ENDWHILE
\ENDFOR
\end{algorithmic}
\end{algorithm}






\bibliographystyle{IEEEtran}
\bibliography{root}

@article{qiu2025biomars,
  title={Biomars: A multi-agent robotic system for autonomous biological experiments},
  author={Qiu, Yibo and Huang, Zan and Wang, Zhiyu and Liu, Handi and Qiao, Yiling and Hu, Yifeng and Sun, Shu'ang and Peng, Hangke and Xu, Ronald X and Sun, Mingzhai},
  journal={arXiv preprint arXiv:2507.01485},
  year={2025}
}

@article{chen2025tevir,
  title={TeViR: Text-to-Video Reward with Diffusion Models for Efficient Reinforcement Learning},
  author={Chen, Yuhui and Li, Haoran and Jiang, Zhennan and Wen, Haowei and Zhao, Dongbin},
  journal={arXiv preprint arXiv:2505.19769v2},
  year={2025},
  note={Submitted to IEEE Transactions on Robotics}
}

@article{chen2025conrft,
  title={Conrft: A reinforced fine-tuning method for vla models via consistency policy},
  author={Chen, Yuhui and Tian, Shuai and Liu, Shugao and Zhou, Yingting and Li, Haoran and Zhao, Dongbin},
  journal={arXiv preprint arXiv:2502.05450},
  year={2025}
}

@inproceedings{kelly2019hg,
  title={Hg-dagger: Interactive imitation learning with human experts},
  author={Kelly, Michael and Sidrane, Chelsea and Driggs-Campbell, Katherine and Kochenderfer, Mykel J},
  booktitle={2019 International Conference on Robotics and Automation (ICRA)},
  pages={8077--8083},
  year={2019},
  organization={IEEE}
}

@article{luo2025precise,
  title={Precise and dexterous robotic manipulation via human-in-the-loop reinforcement learning},
  author={Luo, Jianlan and Xu, Charles and Wu, Jeffrey and Levine, Sergey},
  journal={Science Robotics},
  volume={10},
  number={105},
  pages={eads5033},
  year={2025},
  publisher={American Association for the Advancement of Science}
}

@article{angers2025roboculture,
  title={RoboCulture: a robotics platform for automated biological experimentation},
  author={Angers, Kevin and Darvish, Kourosh and Yoshikawa, Naruki and Okhovatian, Sargol and Bannerman, Dawn and Yakavets, Ilya and Shkurti, Florian and Aspuru-Guzik, Al{\'a}n and Radisic, Milica},
  journal={arXiv preprint arXiv:2505.14941},
  year={2025}
}

@inproceedings{luo2024serl,
  title={Serl: A software suite for sample-efficient robotic reinforcement learning},
  author={Luo, Jianlan and Hu, Zheyuan and Xu, Charles and Tan, You Liang and Berg, Jacob and Sharma, Archit and Schaal, Stefan and Finn, Chelsea and Gupta, Abhishek and Levine, Sergey},
  booktitle={2024 IEEE International Conference on Robotics and Automation (ICRA)},
  pages={16961--16969},
  year={2024},
  organization={IEEE}
}

@article{team2024octo,
  title={Octo: An open-source generalist robot policy},
  author={Team, Octo Model and Ghosh, Dibya and Walke, Homer and Pertsch, Karl and Black, Kevin and Mees, Oier and Dasari, Sudeep and Hejna, Joey and Kreiman, Tobias and Xu, Charles and others},
  journal={arXiv preprint arXiv:2405.12213},
  year={2024}
}

@article{chen2023boosting,
  title={Boosting continuous control with consistency policy},
  author={Chen, Yuhui and Li, Haoran and Zhao, Dongbin},
  journal={arXiv preprint arXiv:2310.06343},
  year={2023}
}

@article{song2023consistency,
  title={Consistency models},
  author={Song, Yang and Dhariwal, Prafulla and Chen, Mark and Sutskever, Ilya},
  year={2023}
}

@article{karras2022elucidating,
  title={Elucidating the design space of diffusion-based generative models},
  author={Karras, Tero and Aittala, Miika and Aila, Timo and Laine, Samuli},
  journal={Advances in neural information processing systems},
  volume={35},
  pages={26565--26577},
  year={2022}
}

@article{salazar2026adept,
  title={The ADePT framework for assessing autonomous laboratory robotics},
  author={Salazar-Villacis, Pablo and Benyahia, Brahim},
  journal={Communications Chemistry},
  volume={9},
  number={1},
  pages={99},
  year={2026},
  publisher={Nature Publishing Group UK London}
}

@article{wolf2025integration,
  title={Integration framework for robotics in life science laboratories Laboratory Automation Plug and Play},
  author={Wolf, {\'A}d{\'a}m},
  year={2025}
}

@article{torres2022automated,
  title={Automated liquid-handling operations for robust, resilient, and efficient bio-based laboratory practices},
  author={Torres-Acosta, Mario A and Lye, Gary J and Dikicioglu, Duygu},
  journal={Biochemical Engineering Journal},
  volume={188},
  pages={108713},
  year={2022},
  publisher={Elsevier}
}

@article{omair2023recent,
  title={Recent advancements in laboratory automation technology and their impact on scientific research and laboratory procedures},
  author={Omair, Abdullah Omar Mohammed and Jabbar, Abeer Mohammed Abdul and Albulushi, Mustafa Othman},
  journal={International journal of health sciences},
  volume={7},
  number={S1},
  pages={3043--3052},
  year={2023},
  publisher={ScienceScholar}
}

@article{thurow2023strategies,
  title={Strategies for automating analytical and bioanalytical laboratories},
  author={Thurow, Kerstin},
  journal={Analytical and Bioanalytical Chemistry},
  volume={415},
  number={21},
  pages={5057--5066},
  year={2023},
  publisher={Springer}
}

@article{hamm2024modular,
  title={A modular robotic platform for biological research: cell culture automation and remote experimentation},
  author={Hamm, Jungmin and Lim, Seonghyeon and Park, Jiae and Kang, Jiwon and Lee, Injun and Lee, Yoongeun and Kang, Jiseok and Jo, Youngjun and Lee, Jaejin and Lee, Seoyeong and others},
  journal={Advanced Intelligent Systems},
  volume={6},
  number={5},
  pages={2300566},
  year={2024},
  publisher={Wiley Online Library}
}

@inproceedings{zhao2025autonomous,
  title={An Autonomous Intelligent Robot for Reagent Handling in Biological Laboratories},
  author={Zhao, Huizhou and Meng, Fansheng and Li, Changsheng and Huan, Jiale and Wang, Xujia and Duan, Xingguang},
  booktitle={2025 IEEE/ASME International Conference on Advanced Intelligent Mechatronics (AIM)},
  pages={1--6},
  year={2025},
  organization={IEEE}
}

@article{seifrid2022autonomous,
  title={Autonomous chemical experiments: Challenges and perspectives on establishing a self-driving lab},
  author={Seifrid, Martin and Pollice, Robert and Aguilar-Granda, Andres and Morgan Chan, Zamyla and Hotta, Kazuhiro and Ser, Cher Tian and Vestfrid, Jenya and Wu, Tony C and Aspuru-Guzik, Alan},
  journal={Accounts of Chemical Research},
  volume={55},
  number={17},
  pages={2454--2466},
  year={2022},
  publisher={ACS Publications}
}

@article{lu2024automated,
  title={Automated Intelligent Platforms for High-Throughput Chemical Synthesis},
  author={Lu, Jia-Min and Pan, Jian-Zhang and Mo, Yi-Ming and Fang, Qun},
  journal={Artificial Intelligence Chemistry},
  volume={2},
  number={1},
  pages={100057},
  year={2024},
  publisher={Elsevier}
}

@article{dai2024autonomous,
  title={Autonomous mobile robots for exploratory synthetic chemistry},
  author={Dai, Tianwei and Vijayakrishnan, Sriram and Szczypi{\'n}ski, Filip T and Ayme, Jean-Fran{\c{c}}ois and Simaei, Ehsan and Fellowes, Thomas and Clowes, Rob and Kotopanov, Lyubomir and Shields, Caitlin E and Zhou, Zhengxue and others},
  journal={Nature},
  volume={635},
  number={8040},
  pages={890--897},
  year={2024},
  publisher={Nature Publishing Group UK London}
}

@article{zhang2025pure,
  title={Pure vision language action (vla) models: A comprehensive survey},
  author={Zhang, Dapeng and Sun, Jing and Hu, Chenghui and Wu, Xiaoyan and Yuan, Zhenlong and Zhou, Rui and Shen, Fei and Zhou, Qingguo},
  journal={arXiv preprint arXiv:2509.19012},
  year={2025}
}

@inproceedings{guo2025improving,
  title={Improving vision-language-action model with online reinforcement learning},
  author={Guo, Yanjiang and Zhang, Jianke and Chen, Xiaoyu and Ji, Xiang and Wang, Yen-Jen and Hu, Yucheng and Chen, Jianyu},
  booktitle={2025 IEEE International Conference on Robotics and Automation (ICRA)},
  pages={15665--15672},
  year={2025},
  organization={IEEE}
}

@article{huang2025co,
  title={Co-rft: Efficient fine-tuning of vision-language-action models through chunked offline reinforcement learning},
  author={Huang, Dongchi and Fang, Zhirui and Zhang, Tianle and Li, Yihang and Zhao, Lin and Xia, Chunhe},
  journal={arXiv preprint arXiv:2508.02219},
  year={2025}
}

@article{deng2025reward,
  title={Reward shaping in reinforcement learning for robotic hand manipulation},
  author={Deng, Zelin and Dong, Yunlong and Liu, Xing},
  journal={Neurocomputing},
  volume={638},
  pages={130204},
  year={2025},
  publisher={Elsevier}
}

@article{fareh2025physics,
  title={Physics-informed reward shaped reinforcement learning control of a robot manipulator},
  author={Fareh, Raouf and Siddique, Tanjulee and Choutri, Kheireddine and Dylov, Dmitry V},
  journal={Ain Shams Engineering Journal},
  volume={16},
  number={10},
  pages={103595},
  year={2025},
  publisher={Elsevier}
}

@article{ibrahim2024comprehensive,
  title={Comprehensive overview of reward engineering and shaping in advancing reinforcement learning applications},
  author={Ibrahim, Sinan and Mostafa, Mostafa and Jnadi, Ali and Salloum, Hadi and Osinenko, Pavel},
  journal={IEEE Access},
  volume={12},
  pages={175473--175500},
  year={2024},
  publisher={IEEE}
}

@article{he2023robotic,
  title={Robotic control in adversarial and sparse reward environments: A robust goal-conditioned reinforcement learning approach},
  author={He, Xiangkun and Lv, Chen},
  journal={IEEE Transactions on Artificial Intelligence},
  volume={5},
  number={1},
  pages={244--253},
  year={2023},
  publisher={IEEE}
}

@article{intelligence2025pi,
  title={$\pi^{*}_{0.6}$: a VLA That Learns From Experience},
  author={Intelligence, Physical and Amin, Ali and Aniceto, Raquel and Balakrishna, Ashwin and Black, Kevin and Conley, Ken and Connors, Grace and Darpinian, James and Dhabalia, Karan and DiCarlo, Jared and others},
  journal={arXiv preprint arXiv:2511.14759},
  year={2025}
}

@article{xu2024rldg,
  title={Rldg: Robotic generalist policy distillation via reinforcement learning},
  author={Xu, Charles and Li, Qiyang and Luo, Jianlan and Levine, Sergey},
  journal={arXiv preprint arXiv:2412.09858},
  year={2024}
}

@article{jin2025dual,
  title={Dual-Actor Fine-Tuning of VLA Models: A Talk-and-Tweak Human-in-the-Loop Approach},
  author={Jin, Piaopiao and Wang, Qi and Sun, Guokang and Cai, Ziwen and He, Pinjia and You, Yangwei},
  journal={arXiv preprint arXiv:2509.13774},
  year={2025}
}

@article{zhang2025reinbot,
  title={Reinbot: Amplifying robot visual-language manipulation with reinforcement learning},
  author={Zhang, Hongyin and Zhuang, Zifeng and Zhao, Han and Ding, Pengxiang and Lu, Hongchao and Wang, Donglin},
  journal={arXiv preprint arXiv:2505.07395},
  year={2025}
}

@article{brohan2022rt,
  title={Rt-1: Robotics transformer for real-world control at scale},
  author={Brohan, Anthony and Brown, Noah and Carbajal, Justice and Chebotar, Yevgen and Dabis, Joseph and Finn, Chelsea and Gopalakrishnan, Keerthana and Hausman, Karol and Herzog, Alex and Hsu, Jasmine and others},
  journal={arXiv preprint arXiv:2212.06817},
  year={2022}
}

@article{kim2024openvla,
  title={Openvla: An open-source vision-language-action model},
  author={Kim, Moo Jin and Pertsch, Karl and Karamcheti, Siddharth and Xiao, Ted and Balakrishna, Ashwin and Nair, Suraj and Rafailov, Rafael and Foster, Ethan and Lam, Grace and Sanketi, Pannag and others},
  journal={arXiv preprint arXiv:2406.09246},
  year={2024}
}

@article{black2024pi_0,
  title={$\pi_0$: A Vision-Language-Action Flow Model for General Robot Control},
  author={Black, Kevin and Brown, Noah and Driess, Danny and Esmail, Adnan and Equi, Michael and Finn, Chelsea and Fusai, Niccolo and Groom, Lachy and Hausman, Karol and Ichter, Brian and others},
  journal={arXiv preprint arXiv:2410.24164},
  year={2024}
}

@inproceedings{esser2021taming,
  title={Taming transformers for high-resolution image synthesis},
  author={Esser, Patrick and Rombach, Robin and Ommer, Bjorn},
  booktitle={Proceedings of the IEEE/CVF conference on computer vision and pattern recognition},
  pages={12873--12883},
  year={2021}
}

@article{yu2023language,
  title={Language to rewards for robotic skill synthesis},
  author={Yu, Wenhao and Gileadi, Nimrod and Fu, Chuyuan and Kirmani, Sean and Lee, Kuang-Huei and Arenas, Montse Gonzalez and Chiang, Hao-Tien Lewis and Erez, Tom and Hasenclever, Leonard and Humplik, Jan and others},
  journal={arXiv preprint arXiv:2306.08647},
  year={2023}
}

@article{wang2023robogen,
  title={Robogen: Towards unleashing infinite data for automated robot learning via generative simulation},
  author={Wang, Yufei and Xian, Zhou and Chen, Feng and Wang, Tsun-Hsuan and Wang, Yian and Fragkiadaki, Katerina and Erickson, Zackory and Held, David and Gan, Chuang},
  journal={arXiv preprint arXiv:2311.01455},
  year={2023}
}

@article{ma2023eureka,
  title={Eureka: Human-level reward design via coding large language models},
  author={Ma, Yecheng Jason and Liang, William and Wang, Guanzhi and Huang, De-An and Bastani, Osbert and Jayaraman, Dinesh and Zhu, Yuke and Fan, Linxi and Anandkumar, Anima},
  journal={arXiv preprint arXiv:2310.12931},
  year={2023}
}

@article{rocamonde2023vision,
  title={Vision-language models are zero-shot reward models for reinforcement learning},
  author={Rocamonde, Juan and Montesinos, Victoriano and Nava, Elvis and Perez, Ethan and Lindner, David},
  journal={arXiv preprint arXiv:2310.12921},
  year={2023}
}

\end{CJK*}
\end{document}